\documentclass{article}
\usepackage{spconf,amsmath,graphicx,hyperref}
\usepackage{tcolorbox}
\usepackage{booktabs} 
\usepackage{multirow}

\title{No Concept Left Behind: Test-Time Optimization for Compositional Text-to-Image Generation}
\name{\begin{tabular}{c}
Mohammad Hossein Sameti\textsuperscript{*}, Amir M. Mansourian\textsuperscript{*}, Arash Marioriyad, Soheil Fadaee Oshyani,\\
Mohammad Hossein Rohban, Mahdieh Soleymani Baghshah
\end{tabular}}

\address{Sharif University of Technology, Tehran, Iran \\
\textsuperscript{*} Equal contribution}

\begin{document}
%
\maketitle
\begin{abstract}
Despite recent advances in text-to-image (T2I) models, they often fail to faithfully render all elements of complex prompts, frequently omitting or misrepresenting specific objects and attributes. Test-time optimization has emerged as a promising approach to address this limitation by refining generation without the need for retraining. In this paper, we propose a fine-grained test-time optimization framework that enhances compositional faithfulness in T2I generation. Unlike most of prior approaches that rely solely on a global image–text similarity score, our method decomposes the input prompt into semantic concepts and evaluates alignment at both the global and concept levels. A fine-grained variant of CLIP is used to compute concept-level correspondence, producing detailed feedback on missing or inaccurate concepts. This feedback is fed into an iterative prompt refinement loop, enabling the large language model to propose improved prompts. Experiments on DrawBench and CompBench prompts demonstrate that our method significantly improves concept coverage and human-judged faithfulness over both standard test-time optimization and the base T2I model. Code is available at: \hyperlink{https://github.com/AmirMansurian/NoConceptLeftBehind}{https://github.com/AmirMansurian/NoConceptLeftBehind}
\end{abstract}
\begin{keywords}
Text-to-Image Generation, Test-time Optimization, Compositionality
\end{keywords}
\section{Introduction}
\label{sec:intro}

\begin{figure}[t]
    \begin{minipage}[t]{0.48\textwidth}
        \hfill 
        \includegraphics[width=\linewidth]{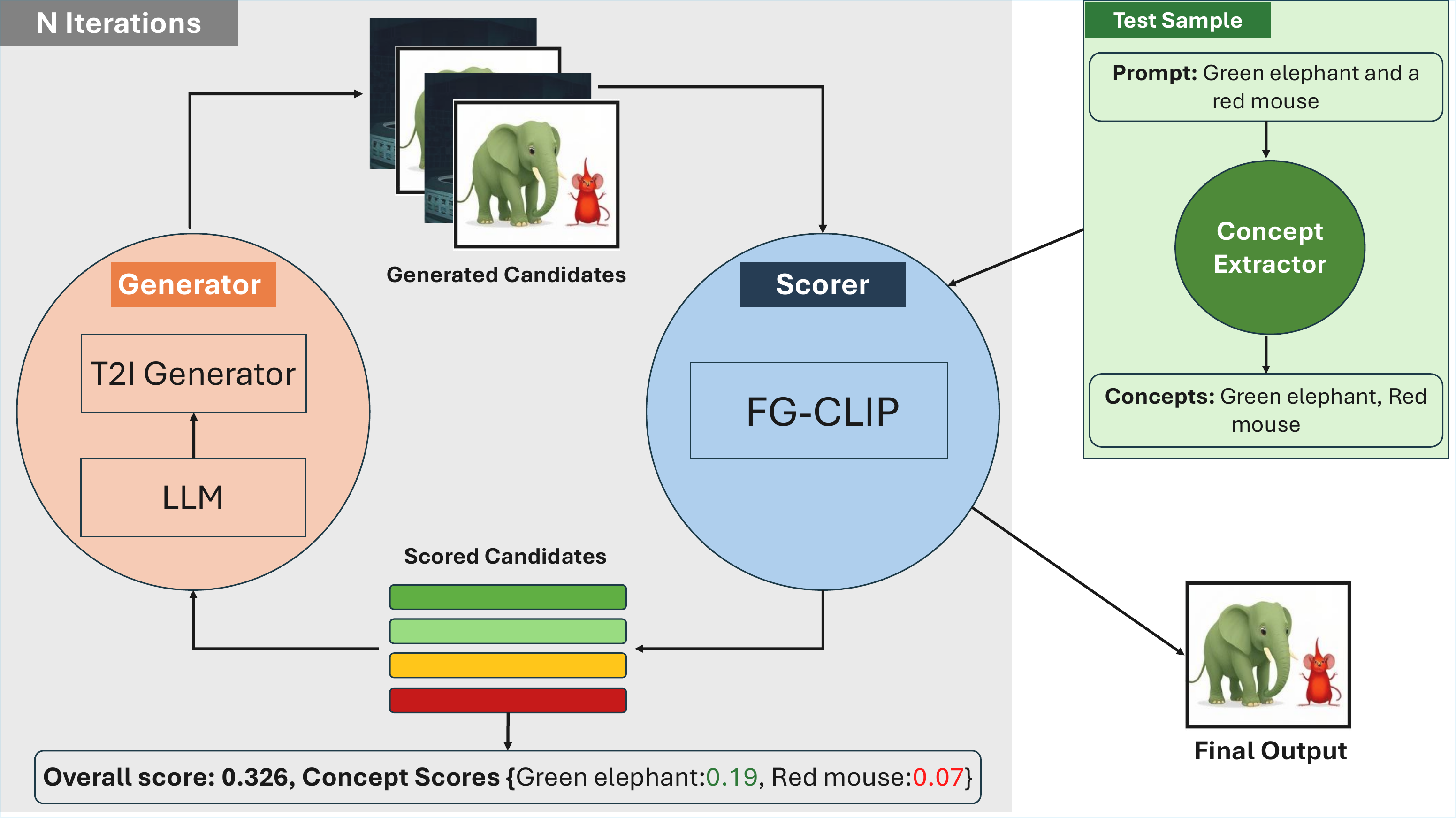}
        \caption{\textbf{Overall diagram of the proposed framework.} Input prompt is first processed by the \textbf{Concept Extractor} module, which extracts key concepts from the initial prompt. Then, in each step, the \textbf{Generator} module uses an LLM to rewrite diverse prompts based on both overall and concept-level scores, and a T2I model generates candidate images. The \textbf{Scorer} module evaluates these candidates using a fine-grained variant of CLIP, and the scores are fed back to the generator. After a certain number of iterations, the final image is produced. Concept-level scores help the LLM address missing concepts from previous steps and rewrite improved prompts.}
        
        \label{fig:diagram}
    \end{minipage}
\end{figure}

Text-to-image (T2I) generation has seen rapid progress with models such as DALL·E~\cite{ramesh2022hierarchical}, Stable Diffusion~\cite{rombach2022high}, and FLUX~\cite{flux2024, batifol2025flux}, which can synthesize realistic images from natural language descriptions. Despite impressive zero-shot capabilities, these models often struggle with compositional faithfulness: faithfully representing all objects, attributes, and relations described in the prompt~\cite{qu2025silmm, wan2025compalign}. 

A promising direction to mitigate such failures is test-time optimization. Instead of retraining large models, one can refine the input prompt or generation process iteratively, guided by a scoring function. Recent frameworks demonstrate that large language models (LLMs) can be used at inference time to generate and refine candidate prompts, while an external scorer evaluates image–text alignment~\cite{singh2023divide, mrini2024fast, khan2025test, ashutosh2025llms}. This loop can progressively improve results without additional training data. One such framework is MILS \cite{ashutosh2025llms} (Multimodal Iterative LLM Solver), which leverages an LLM to propose prompts and a global CLIP-based~\cite{radford2021learning} scorer to evaluate generated images. While MILS demonstrates the potential of test-time iterative optimization, it is limited by its reliance on a single, coarse similarity score. A global CLIP similarity may assign a high score even if some prompt concepts are missing, since it measures overall match rather than concept-level fidelity.

\textbf{Contributions.} 
We propose a fine-grained test-time optimization framework for text-to-image generation that improves compositional fidelity (Figure~\ref{fig:diagram}). 
Our approach extends iterative prompt refinement by introducing concept extraction and a fine-grained CLIP scorer that evaluates both global prompt–image alignment and per-concept correspondence. 
By feeding these detailed scores back to the LLM, the system is guided to explicitly recover missing objects, attributes, and relations, resulting in images that more faithfully capture all aspects of the input prompt while maintaining overall semantic alignment.


\section{Related Work}
\label{sec:related}

T2I generation has rapidly advanced from early GAN-based approaches to transformer-based and diffusion models that achieve striking visual quality~\cite{xu2018attngan, zhu2019dm, ramesh2022hierarchical, saharia2022photorealistic, rombach2022high}. 
Recent large-scale diffusion models, including DALL·E~2~\cite{ramesh2022hierarchical}, Stable Diffusion~\cite{rombach2022high}, and FLUX~\cite{flux2024} demonstrate impressive zero-shot generation capabilities across diverse prompts. Despite these advances, even state-of-the-art models remain prone to compositional errors, often omitting or misrepresenting specific objects, attributes, or spatial relations when prompts become complex or multi-object in nature.

One promising direction for addressing this challenge is test-time optimization, which adapts or refines model behavior during inference without additional training, often improving robustness and alignment with task-specific requirements. Several recent studies have demonstrated that prompt optimization at test time can significantly enhance performance. For instance, ~\cite{singh2023divide} iteratively rewrites prompts by obtaining feedback from a visual question answering model, another variant refines prompts using a multimodal large language model~\cite{khan2025test}, and MILS~\cite{ashutosh2025llms} introduces an iterative framework driven by CLIP-based feedback. To reduce the inference-time burden of such methods,~\cite{mrini2024fast} also proposes a fast, single-iteration prompt alignment.


\begin{table*}[th!]
\centering
\setlength{\tabcolsep}{6pt} 
\renewcommand{\arraystretch}{1.1} 
\caption{Quantitative results comparison on T2I CompBench and DrawBench prompts.}
\label{tab:unified_t2i_drawbench}
\begin{tabular}{lcccccccc}
\toprule
\multirow{2}{*}{\textbf{Method}} 
& \multicolumn{4}{c}{\textbf{T2I CompBench}} 
& \multicolumn{4}{c}{\textbf{DrawBench}} \\
\cmidrule(lr){2-5} \cmidrule(lr){6-9}
& \textbf{VQA} & \textbf{CLIP (L)} & \textbf{Captioning (L)} & \textbf{GPT4-o} 
& \textbf{VQA} & \textbf{CLIP (L)} & \textbf{Captioning (L)} & \textbf{GPT4-o} \\
\midrule
FLUX & 0.865 & 0.272 & 0.687 & 0.717 & 0.620 & 0.279 & 0.645 & 0.719 \\
MILS & 0.925 & 0.287 & 0.694 & 0.744 & 0.665 & 0.299 & 0.671 & 0.765\\
Our  & \textbf{0.955} & \textbf{0.295} & \textbf{0.701} & \textbf{0.810} 
      & \textbf{0.715} & \textbf{0.304} & \textbf{0.677} & \textbf{0.827} \\
\bottomrule
\end{tabular}
\end{table*}

\section{Method}
\label{sec:method}

\subsection{Overview}
Given an input prompt $P$, we first extract its key semantic concepts (e.g., objects, attributes, relations), and then evaluate generated images with both a global similarity score and per-concept scores computed via a fine-grained CLIP model. This feedback is used in an iterative loop to refine candidate prompts proposed by an LLM, leading to images that more faithfully capture all prompt elements. Figure \ref{fig:diagram} illustrates the overall diagram of the proposed method.

\subsection{Concept Extraction}
Let $P$ be the input prompt, decomposed into a set of $k$ concepts: $\mathcal{C}(P) = \{c_1, c_2, \dots, c_k\}$.
Concepts include objects, attributes, and relations.
We obtain $\mathcal{C}(P)$ using a syntactic parser or an LLM-based semantic extractor. 

\subsection{Fine-Grained Scoring}
Given the input prompt ${P}$ and the generated image $I$ 
from the T2I model, we compute two types of similarity scores:

\begin{align}
    S_{\text{global}}(I, {P}) &= \text{CLIP}(I, {P}), \\
    s_i(I, c_i) &= \text{CLIP}(I, c_i), \quad i = 1,\dots,k,
\end{align}

\noindent where $\text{CLIP}(\cdot,\cdot)$ denotes cosine similarity in the joint embedding space of a fine-grained CLIP variant.
The global score measures alignment with the entire prompt, while $s_i$ evaluates the presence of each concept individually.

\subsection{Optimization Formulation}
The standard MILS~\cite{ashutosh2025llms} framework optimizes only the global score:
\begin{align}
    \max_{I} \; S_{\text{global}}(I, {P}).
\end{align}
In contrast, we formulate the problem as a multi-objective optimization:

\begin{align}
    \max_{I} \; 
     \, S_{\text{global}}(I, P) 
    + \frac{1}{k}\sum_{i=1}^k s_i(I, c_i),
\end{align}

\noindent where this objective explicitly encourages generated images to satisfy all extracted concepts.

\subsection{Iterative Refinement}
At each iteration, the LLM proposes a batch of candidate prompts. 
For each candidate $\tilde{P}$, images are generated and scored using the above objective.
The top-ranked candidates, along with detailed per-concept scores, are fed back into the LLM
to guide subsequent generations. 
This loop continues until convergence or a fixed number of iterations.

\subsection{Signal Processing Viewpoint}
Our method can be interpreted through the lens of signal processing. The prompt $P$ is analogous to a composite signal with semantic ``frequency components'' $\{c_i\}$. A global similarity score reflects total signal energy, which may remain high even if some bands are missing. By projecting the generated image $I$ onto each concept $c_i$ via $s_i(I, c_i)$, we perform a filter-bank--like decomposition. Iteratively refining prompts with these projections resembles adaptive equalization, ensuring all components are preserved. This perspective grounds our method: it enforces both global alignment and the faithful preservation of individual concepts, akin to maintaining overall energy and spectral detail in classical signal reconstruction.

\begin{figure}[ht!]
    \begin{minipage}[t]{0.48\textwidth}
        \hfill 
        \includegraphics[width=\linewidth]{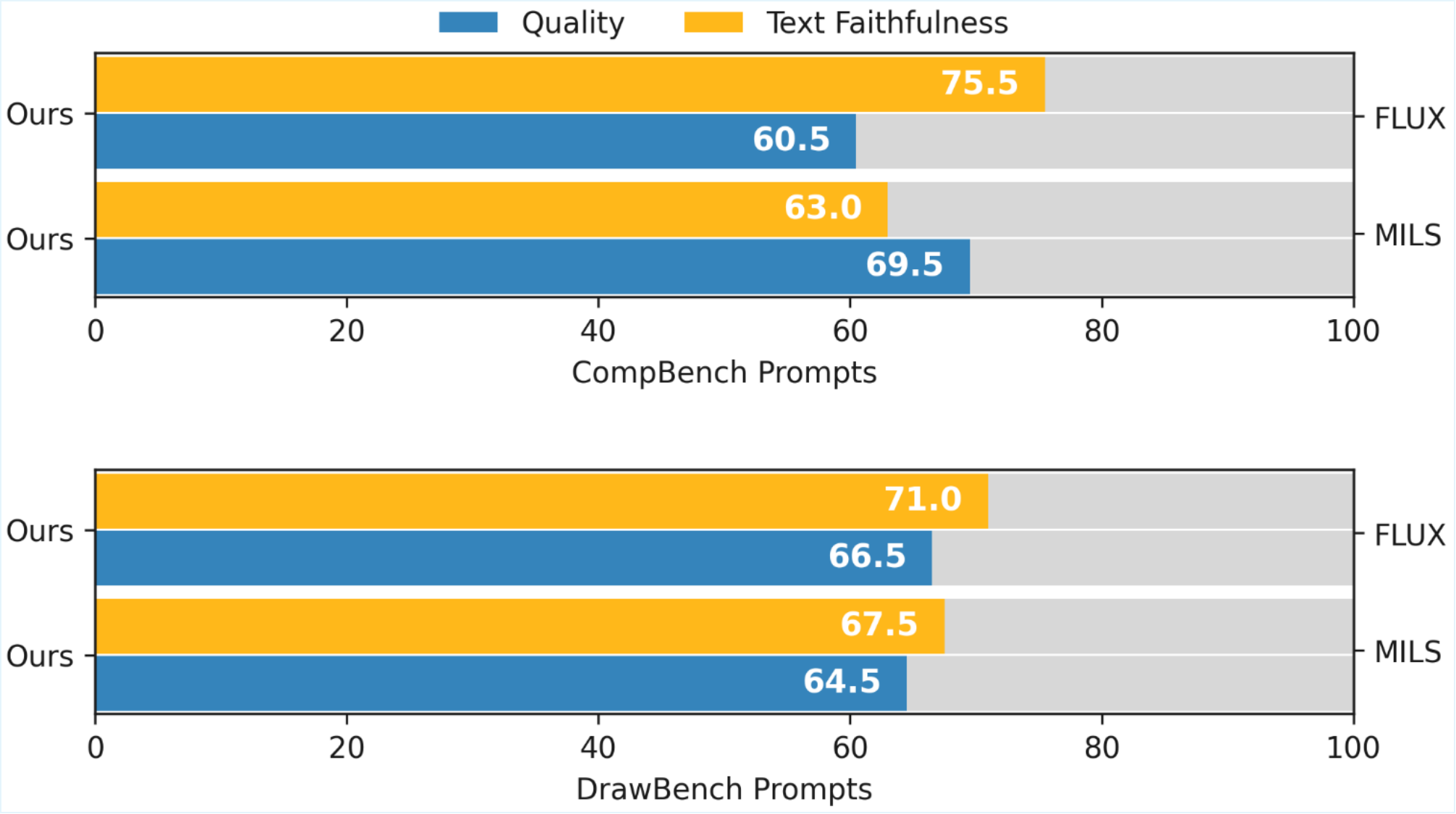}
        \caption{Win rate comparison judged by human evaluation.}
        \label{fig:winrate}
    \end{minipage}
\end{figure}

\begin{figure*}[ht!]
	\centerline{\includegraphics[width=0.95\textwidth]{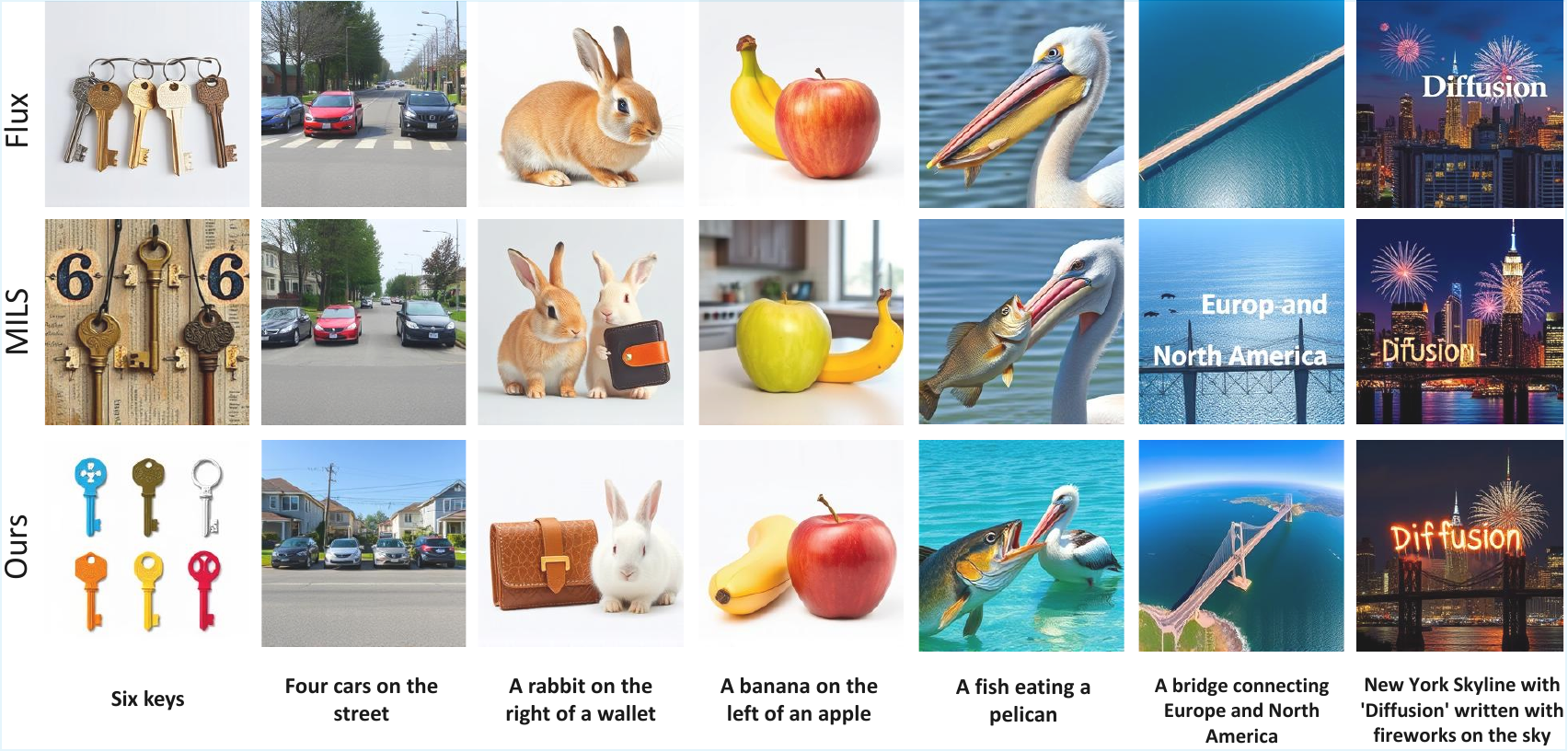}}
	\caption{Qualitative comparison with FLUX and MILS baselines on sample prompts from DrawBench and CompBench across different categories.}
	\label{fig:qual_results}
\end{figure*}

\section{Experiments}
\label{sec:experiment}


\subsection{Datasets and Evaluation Metrics}
We evaluate our method using two widely adopted benchmarks. First, DrawBench \cite{saharia2022photorealistic}, a set of 200 prompts designed to test text-image alignment, across 11 categories. Second, a curated subset of T2I-CompBench \cite{huang2023t2i}, where we select 200 prompts distributed across 8 compositional categories. Details about the categories and sampling procedure are provided in Appendix~\ref{appendix:datasets}.


For evaluation, we employ both human and automated metrics. Human evaluation follows a win-rate protocol against the baseline under clear guidelines to reduce subjectivity, with three evaluators whose judgments are aggregated by majority vote. Automated evaluation consists of several complementary metrics: Visual Question Answering (VQA) using BLIP \cite{li2022blip}, CLIP Score \cite{hessel2021clipscore} measuring text–image similarity in the CLIP embedding space, and Captioning Score, which computes the similarity between the BLIP-generated caption and the original prompt in the CLIP text embedding space. We further introduce a GPT Score, where GPT-4o rates alignment between prompt and image on a $[0,1]$ scale, using category-specific prompts (e.g., attributes, spatial relations, numeracy) to capture different types of compositional alignment. Details about the evaluation metrics are provided in Appendix~\ref{appendix:evaluation_metrics}.

\begin{figure}[t]
    \begin{minipage}[t]{0.48\textwidth}
        \hfill 
        \includegraphics[width=\linewidth]{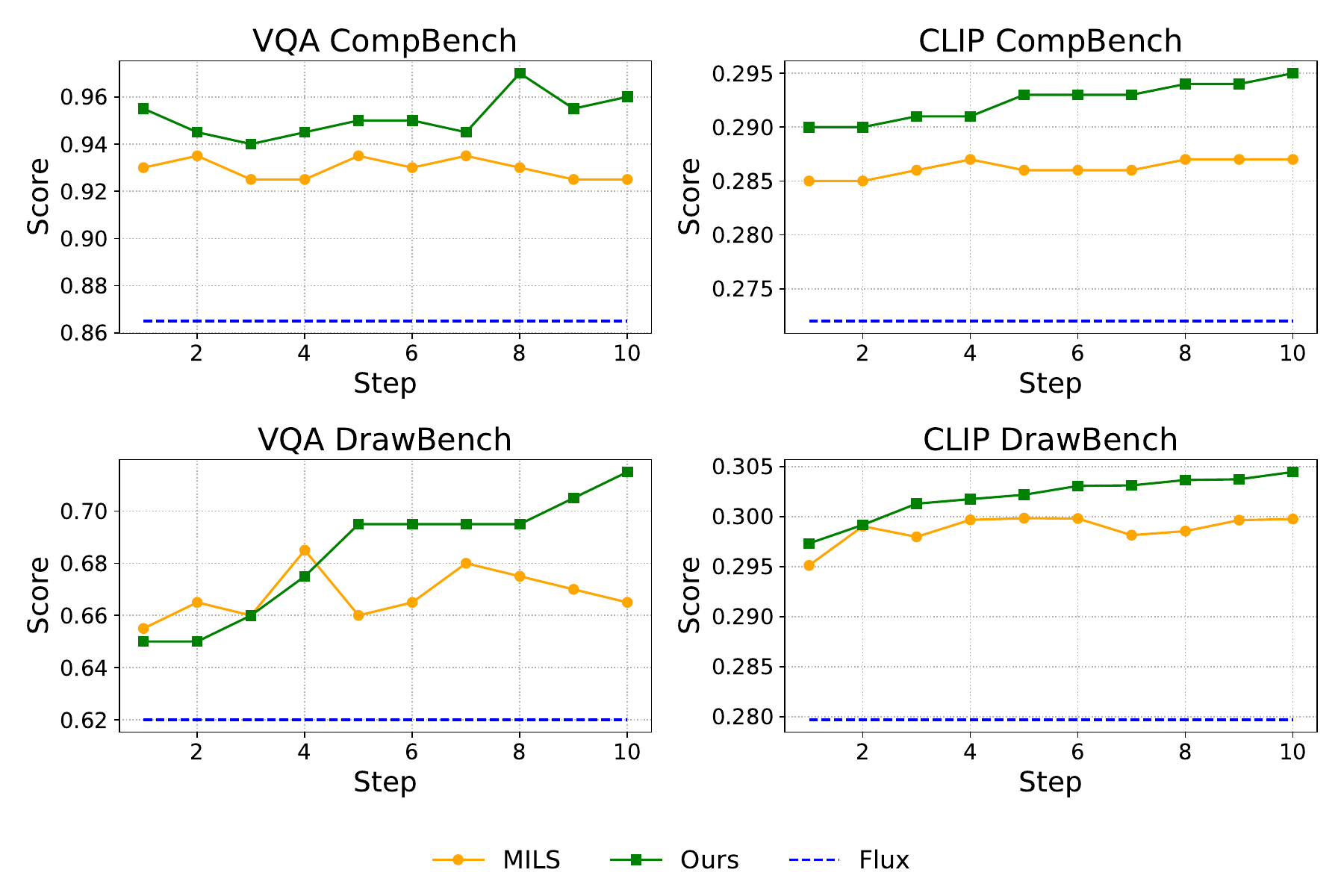}
        \caption{Per-step comparison of our method with Flux and MILS baselines across two metrics on the DrawBench and CompBench prompts.}
        \label{fig:step}
    \end{minipage}
\end{figure}

\subsection{Implementation Details}
We extend the MILS~\cite{ashutosh2025llms} codebase by integrating several key components. As the scorer, we employ FG-CLIP~\cite{xie2025fg}, a fine-grained variant of CLIP trained for region-to-word alignment. For text-to-image generation, we adopt FLUX.1 [schnell]~\cite{flux2024}, and for re-writing prompts and extracting concepts from the input, we utilize LLaMA-3.1-8B-Instruct~\cite{dubey2024llama}.

During optimization, 50 new prompts are generated at each iteration, while the top 20 prompts from the previous step are retained. The process runs for a total of 10 iterations. Prompt templates for re-writing and concept extraction are provided in the Appendix~\ref{appendix:prompts}.

\subsection{Quantitative and Qualitative Results}
Table~\ref{tab:unified_t2i_drawbench} presents the evaluation metrics for the base generator method (FLUX), the MILS baseline, and our proposed method on two datasets. As shown in the table, our method consistently outperforms the baselines across all metrics, demonstrating the effectiveness of incorporating fine-grained views to enhance compositional faithfulness in T2I generation. In particular, our method yields significant improvements over the FLUX baseline and achieves a clear margin of superiority over the MILS framework across nearly all metrics.

Furthermore, Figure~\ref{fig:winrate} illustrates the win rate of our method compared with FLUX and MILS in terms of both image quality and text faithfulness. The results confirm that images generated by our method are more frequently preferred in human evaluations.

In addition, Figure~\ref{fig:qual_results} provides a qualitative comparison between our method and the baselines. It can be observed that the proposed method enhances the quality of generated images across various categories, including counting, spatial relationships, text rendering, and conflicting prompts.

\subsection{Ablation Study}
To further validate the effectiveness of the proposed method, we conducted ablation studies. Figure \ref{fig:step} presents the results of our method across different iterations and evaluation metrics on both datasets. It can be observed that even in the first iteration, our method achieves strong performance compared to the FLUX baseline and almost consistently outperforms MILS at every iteration. As noted earlier, test-time optimization methods typically introduce additional computational overhead; however, our method surpasses the baselines even at the first step. This flexibility allows practitioners to obtain strong results with minimal overhead, while further improvements can be achieved by increasing the number of iterations if additional computation is acceptable.

\begin{figure}[t]
    \begin{minipage}[t]{0.48\textwidth}
        \hfill 
        \includegraphics[width=\linewidth]{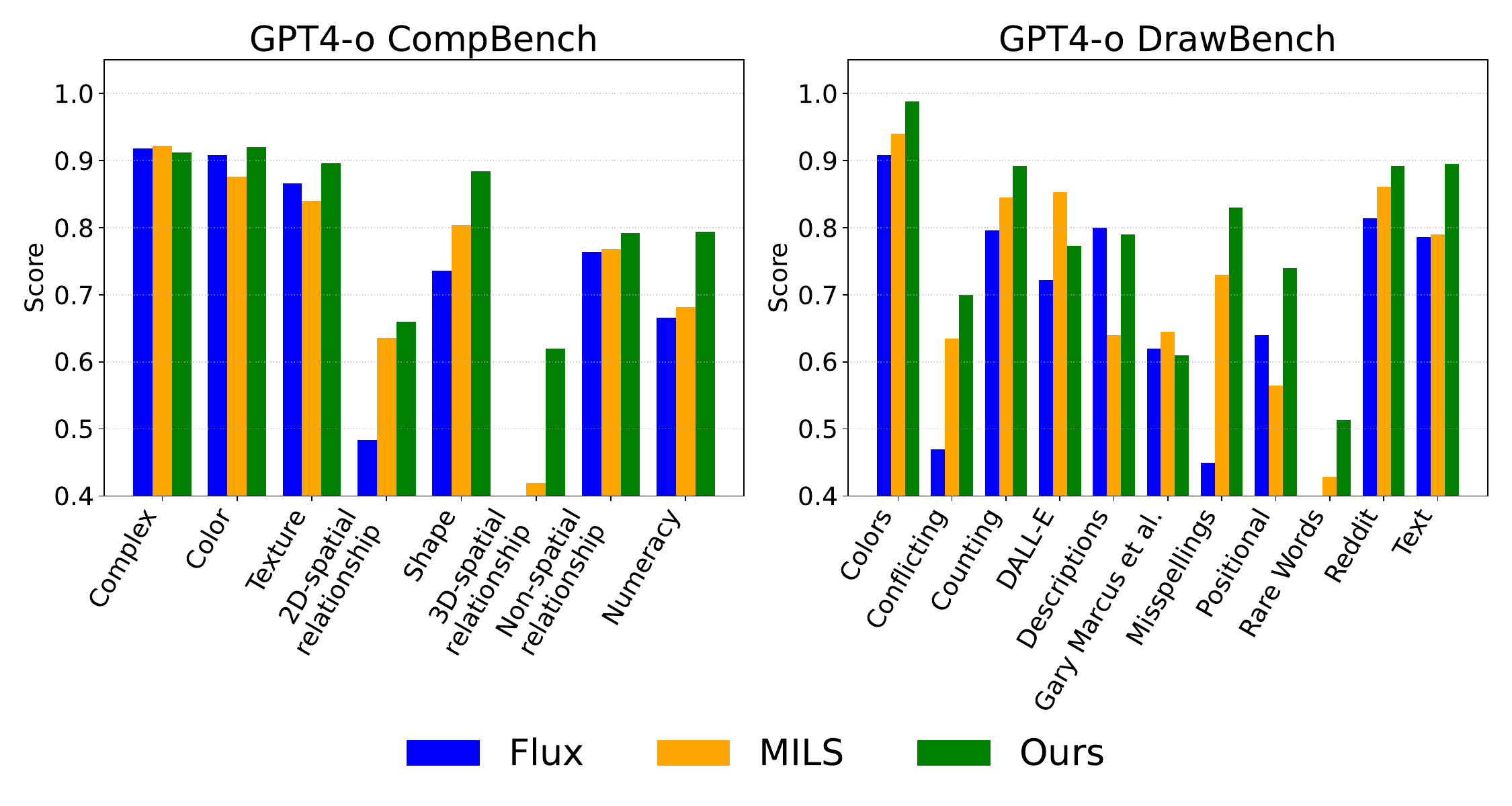}
        \caption{Comparison of GPT4-o scores for the proposed method against FLUX and MILS baselines across different prompt categories on DrawBench and CompBench.}
        \label{fig:category}
    \end{minipage}
\end{figure}

In addition, Figure \ref{fig:category} presents an ablation study of our method’s performance across different prompt categories in comparison to the FLUX and MILS baselines. The results show that our method achieves performance that is consistently better than or on par with MILS across all categories. Notably, it performs marginally better in challenging categories such as counting, spatial relationships, and color prompts, which is consistent with the qualitative comparisons presented earlier in Figure \ref{fig:qual_results}. Thanks to the fine-grained view of our method, compositional improvements are achieved throughout the process, leading to marginally better results in certain categories. Examples of generated images and their corresponding prompts at intermediate iterations are provided in the Appendix~\ref{appendix:examples}.

\section{Discussion}
Fine-grained scoring provides interpretable feedback, explicitly identifying which concepts are underrepresented. This enables the LLM to focus its refinements on the missing elements rather than making generic changes. The result is improved compositional fidelity without sacrificing visual quality. The main limitation lies in concept extraction accuracy and the reliance on CLIP for semantic judgments. Future work could explore combining fine-grained CLIP with object detectors or segmentation models for more robust scoring.

\section{Conclusion}
We presented a fine-grained test-time optimization method for text-to-image generation. By decomposing prompts into concepts and scoring them with a fine-grained CLIP model, our approach supplies detailed feedback to an iterative refinement loop. This yields images that more faithfully capture all aspects of the input prompt, outperforming both MILS and raw T2I generation baselines. Our method highlights the promise of concept-aware test-time optimization for improving the compositionality of generative models.

\bibliographystyle{IEEEbib}
\bibliography{Template}

\appendix
\section{Datasets and Evaluation Metrics}

\subsection{Datasets}
\label{appendix:datasets}

For a comprehensive evaluation, we employ two complementary datasets that test different aspects of compositional text-to-image generation. The first dataset is DrawBench \cite{saharia2022photorealistic}, which consists of 200 prompts carefully constructed to probe a wide spectrum of linguistic and visual phenomena. The prompts are grouped into eleven categories that reflect key challenges in text-to-image generation: ... This dataset is particularly useful because it provides short and interpretable prompts that isolate specific sources of compositional error, enabling controlled comparisons across methods.

In addition to DrawBench, we also evaluate on a curated subset of T2I-CompBench \cite{huang2023t2i}, a benchmark designed to rigorously test compositional generalization. From this benchmark, we select 200 prompts divided evenly across eight categories, with 25 prompts in each. The selected categories cover color binding, texture binding, shape binding, 2D spatial relation, 3D spatial relation, non-spatial relation, numeracy, and complex. This controlled subset allows us to systematically probe distinct forms of compositional reasoning while keeping the evaluation scale manageable and consistent across categories.

\subsection{Evaluation Metrics}
\label{appendix:evaluation_metrics}

To assess the quality of generated images, we adopt both human-based and automated evaluation metrics. Human evaluation follows a win-rate protocol in which three independent annotators compare outputs from our method and the baseline model. For each prompt, the annotators are presented with the text prompt and two corresponding images side by side, without any indication of which image comes from which method. They are asked to judge which image better reflects the compositional alignment with the prompt. We then compute the average win rate across annotators to produce a robust measure of human preference. This setup ensures fairness and minimizes bias by anonymizing the image sources.

\begin{figure}[th]
\centering
\begin{tcolorbox}[colback=gray!5,colframe=black!50,title={Prompt for Concept Extraction}]
\small
\textbf{Task:} \\
Extract multiple essential key or compositional phrases from the image description. 
These phrases should represent:
\begin{itemize}
    \item Visually grounded object descriptions
    \item Compositional variants (e.g., \emph{Big wooden teddy} $\rightarrow$ \emph{big teddy}, \emph{wooden teddy})
    \item Object--Subject--Relation (e.g., \emph{big wooden teddy stands under green tree} $\rightarrow$ \emph{teddy stands under tree})
\end{itemize}

Do not add extra interpretation or paraphrasing. Extract only from the sentence. \\
Generate \textbf{only} extracted tokens in comma-separated format (\emph{e.g. big teddy, wooden teddy, teddy stand under tree, ...})

\medskip
\texttt{\#\#\#image description: \{init\_description\}} \\
\texttt{\#\#\#extracted tokens:}
\end{tcolorbox}
\caption{Prompt template used for extracting concepts.}
\label{fig:prompt_tokenextraction}
\end{figure}

Alongside human evaluation, we employ several automated metrics. Visual Question Answering (VQA) accuracy is measured by applying models such as BLIP \cite{li2022blip}, which answer automatically derived questions about the generated images. CLIP Score \cite{hessel2021clipscore} is computed as the cosine similarity between the CLIP embeddings of the prompt and the generated image, offering a widely used proxy for text–image alignment. To further capture semantic correspondence, we also calculate a Captioning Score: for each generated image, a BLIP captioning model \cite{li2022blip} produces a textual description, which is then compared against the original prompt in the CLIP text embedding space following \cite{hessel2021clipscore}. Finally, to make the evaluation more diverse and robust, we include a GPT-based metric, in which GPT-4o is provided with both the prompt and the generated image and asked to return a scalar score in the range $[0,1]$ that reflects the alignment between the two. This multimodal assessment complements traditional metrics by incorporating reasoning abilities from a state-of-the-art vision-language model.

\begin{figure}[th]
\centering
\begin{tcolorbox}[colback=gray!5,colframe=black!50,title={Prompt for Description Rephrasing}]
\small
You need to expand and rephrase the provided description for image generation to make the best image, by maximizing the image score:

\begin{itemize}
    \item \textbf{Overall score:} how well the full description matches the generated image.
    \item \textbf{Word scores:} how well individual key tokens (objects, attributes, relations) from the description align with the image.
\end{itemize}

\textbf{Current Description:} \\
\{\texttt{init\_description}\} 

\medskip
Here are some example rephrases and the corresponding image scores and token-level alignment scores: \\
\{\texttt{descriptions}\} 

\medskip
\textbf{Instruction:} \\
Generate 50 new descriptions that maximize both the overall and token-level scores. Produce substantive rewrites that strengthen weak tokens and preserve high-scored tokens from prior examples. Place each instruction on a separate line, with a numeric counter (e.g., ``1. ... 2. ...''), and ensure each is concise ($<$77 words).
\end{tcolorbox}
\caption{Prompt template used for generating improved image descriptions.}
\label{fig:prompt_gen}
\end{figure}

\section{Prompts}
\label{appendix:prompts}

In this work, two carefully designed prompt templates form the foundation of our optimization framework. The first, illustrated in Figure~\ref{fig:prompt_tokenextraction}, is dedicated to concept extraction. Its purpose is to decompose an input description into a set of visually grounded and semantically essential units—objects, attributes, and relations—without introducing additional interpretation beyond the original text. This process ensures that every critical element of the description, such as “blue pizza” or “teddy stands under tree,” is explicitly identified as an independent concept. To achieve this, the prompt enforces a structured output format, which provides a reliable basis for fine-grained alignment checks at the level of individual concepts.

The second prompt, shown in Figure\ref{fig:prompt_gen}, is designed for description rephrasing and expansion. At this stage, both global and per-concept scores are leveraged to iteratively refine the input description. The prompt instructs the model to generate multiple rewritten versions of the original description, with the dual goal of maximizing overall semantic alignment and strengthening weaker tokens flagged by concept-level scores. For example, if an object such as a “red mouse” is consistently underrepresented in generated images, the rewritten descriptions place stronger emphasis on that element in subsequent iterations. At the same time, the prompt encourages preservation of concepts already aligned well, thereby ensuring comprehensive coverage without introducing redundancy.

Together, these two prompts create a closed feedback loop. The concept extraction prompt isolates and formalizes the essential building blocks of the description, while the rephrasing prompt integrates global and fine-grained feedback to reinforce underrepresented elements. This iterative interaction allows the optimization process to systematically recover missing details while maintaining fluent and natural phrasing. By alternating between structured decomposition and score-guided rewriting, the framework ensures that generated images not only reflect the overall intent of the prompt but also faithfully capture each constituent concept.

\section{Method Output Examples}
\label{appendix:examples}
\begin{figure*}[ht!]
	\centerline{\includegraphics[width=0.95\textwidth]{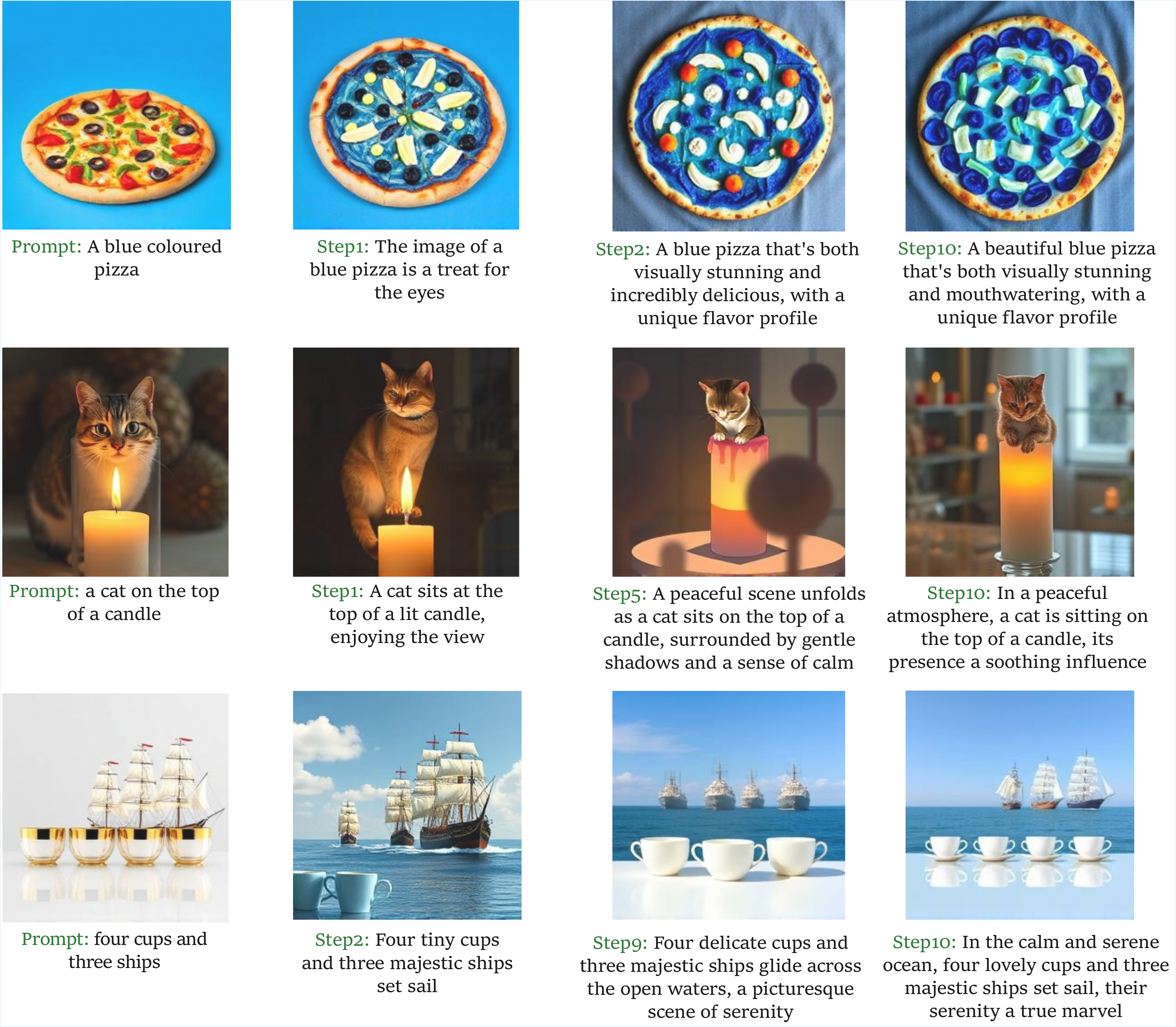}}
	\caption{Improved images are generated at each iteration of the proposed method, along with the corresponding prompts. In each iteration, the prompt is optimized by the LLM, leading to higher-quality output images.}
	\label{fig:steps_output}
\end{figure*}

The iterative nature of our method makes it possible to generate more faithful images by improving the prompt at each iteration of the optimization process. Figure \ref{fig:steps_output} shows the outputs of the method across iterations for several samples in different prompt categories such as coloring, position, and counting. It also presents the corresponding generated prompt for each iteration.

As can be seen, at each iteration the LLM re-generates improved prompts based on feedback from the scorer, which provides both overall and concept-level scores. This guides the LLM to avoid missing any concepts in the prompt.

In the first row, the FLUX model is prompted with “A blue colored pizza”. It initially produces a normal pizza with a blue background. At the first step, however, our method generates the correct image corresponding to the input prompt. In subsequent steps, our method continues to generate a blue pizza, but with progressively better quality. This demonstrates that our method can achieve strong results even in the early iterations, allowing inference time to be reduced by limiting the number of steps.

In the second row, we again observe that the FLUX model fails to generate the correct output. In contrast, our method produces more faithful images at steps 1 and 5, with further improvements in quality by step 10.

In the final example, the method still fails to produce the correct image by the 9th step, as the number of objects is incorrect. However, by the 10th step, our method successfully generates the correct image with good quality. This further validates the effectiveness of the fine-grained feedback incorporated in our proposed method.

\end{document}